# Advancing Complex Medical Communication in Arabic with Sporo AraSum: Surpassing Existing Large Language Models

A comparative study between Sporo Health's AraSum and Inception AI's JAIS.

**Authors:** Chanseo Lee BS,[1,2] Sonu Kumar MTech,[1] Kimon A. Vogt MS,[1] Sam Meraj MBBS,[1] Antonia Vogt[3]

1. Sporo Health, Boston, MA
2. Yale School of Medicine, New Haven, CT
3. Cambridge University, Clinical School, Department of Surgery

## Abstract

The increasing demand for multilingual capabilities in healthcare underscores the need for AI models adept at processing diverse languages, particularly in clinical documentation and decision-making. Arabic, with its complex morphology, syntax, and diglossia, poses unique challenges for natural language processing (NLP) in medical contexts. This case study evaluates Sporo AraSum, a language model tailored for Arabic clinical documentation, against JAIS, the leading Arabic NLP model. Using synthetic datasets and modified PDQI-9 metrics modified ourselves for the purposes of assessing model performances in a different language. The study assessed the models' performance in summarizing patient-physician interactions, focusing on accuracy, comprehensiveness, clinical utility, and linguistic-cultural competence.

Results indicate that Sporo AraSum significantly outperforms JAIS in AI-centric quantitative metrics and all qualitative attributes measured in our modified version of the PDQI-9. AraSum's architecture enables precise and culturally sensitive documentation, addressing the linguistic nuances of Arabic while mitigating risks of AI hallucinations. These findings suggest that Sporo AraSum is better suited to meet the demands of Arabic-speaking healthcare environments, offering a transformative solution for multilingual clinical workflows. Future research should incorporate real-world data to further validate these findings and explore broader integration into healthcare systems.

## Introduction

The increasing demand for multilingual capabilities in healthcare technology highlights a crucial need for AI models capable of effectively processing diverse languages, especially in clinical documentation and decision-making. Medical documentation in a patient's native language can significantly improve understanding, communication, and overall patient care outcomes, particularly in regions where healthcare professionals and patients may not share a common language, or where nuanced medical terminology requires precise understanding. In this regard, Arabic presents unique challenges for language models due to its rich morphological structure, complex syntax, and diglossia,[1] which is the coexistence of formal Arabic and regional dialects. Effective scribing and clinical workflow models that operate seamlessly in Arabic can facilitate accurate documentation, reduce medical errors, and bridge communication gaps that are crucial in patient care.

Arabic poses distinct challenges for AI-based language models that are often trained on clinical information from well-represented Western or Eastern languages, like English or Chinese.[2] Unlike these languages, Arabic is highly inflected, and small changes in root morphology can drastically alter word meanings.[3] In clinical contexts, where precision is paramount, capturing these subtleties can mean the difference between an accurate diagnosis and potential misinterpretation. Research underscores that while natural language processing (NLP) for Arabic has advanced, especially with the introduction of Arabic-specific models like AraBERT[4] and more recently Inception AI's JAIS,[5] the field still lacks models that perform well in specialized medical domains.

Existing Arabic models often struggle with terminology specific to clinical documentation, where meanings can vary greatly depending on context – a problem much exacerbated by the lack of comprehensive medical conversation data that can offer validation.[6] Clinical workflows and documentation require precise, consistent, and interpretable summaries to be effective; for Arabic-speaking patients and providers, this demands a model that is tailored to handle these linguistic intricacies with a focus on medical terminology, as well as an ability to contextualize and summarize clinical data in Arabic without struggling with the language's flexibility.

This case study examines Sporo AraSum, Sporo Health's advanced language model specifically tailored to address these issues in the Arabic domain, and its efficacy in addressing the complexities of clinical documentation, scribing, workflow, and decision-making compared to JAIS, the leading Arabic language model currently available. The study explores how Sporo AraSum's architecture and training are uniquely optimized to handle medical terminology, syntax, and cultural nuances in Arabic, enabling it to better understand and summarize clinical interactions. By comparing Sporo AraSum's performance against JAIS in key areas, the study aims to illustrate Sporo AraSum's potential to elevate the standard of care for Arabic-speaking patients and medical professionals.

## Methods

*Evaluating summary quality using known quantitative metrics*

Although there is a lack of proper medical conversation datasets in Arabic, AlMutairi et al. showed that synthetic data created by LLMs can suffice as a data source.[6] We generated 4,000 synthetic patient-physician conversations in Arabic using GPT-4o, and AI-generated summaries were created using Sporo AraSum and JAIS, then compared to ground truth clinical summaries generated by GPT-4o and translated to Arabic.

**Clinical content recall** was defined as the proportion of relevant clinical information from the ground truth summary that was accurately captured in the AI-generated summaries. Salient clinical items were extracted from each conversation into an inventory. Recall was then calculated by dividing the number of correctly included items from the inventory by the total number of relevant items. **Clinical content precision** was defined as the proportion of information that was both accurate and relevant when compared to the  ground truth. Precision was calculated by dividing the number of correctly included items in the summary by the total number of items, including any additional or incorrect items. This metric reflects the accuracy and relevance of the content without introducing extraneous or inaccurate details. Finally, the **F1 score** is used as a balanced metric to combine both clinical content precision and recall, providing a single measure of the AI-generated summaries' performance.[7] It represents the harmonic mean of precision and recall, ensuring that both the accuracy of relevant information captured (precision) and the completeness of that information (recall) are taken into account. ROUGE scores,[8] BLEU,[9] and BERTScore F1[10] were also calculated.

*Qualitative evaluation of clinical utility*

Three transcripts of patient-physician conversations covering simple clinical vignettes were created in Arabic. The patient presentations for each conversation included a rash caused by new soap, heart palpitations and hypertension, and tonsillitis. AI-generated clinical summaries were generated by Sporo AraSum and JAIS using zero-shot prompting. Each generated clinical summary was compared to a clinician-generated summary translated into Arabic that served as the ground truth.

External evaluators fluent in Arabic were provided the conversation transcripts and the two AI-generated summaries, blinded to the source and consistent in formatting. Summary satisfaction was evaluated in conjunction with accuracy using a modified version of the **Physician Documentation Quality Instrument revision 9 (PDQI-9)**. The original PDQI-9 employs a 5-point Likert scale across nine attributes to assess note quality. This was modified by Tierney et al. into a ten-item inventory to better fit the metrics relevant to ambient AI documentation, and is widely used to evaluate AI-generated clinical notes.[11],[12] Three additional language-specific attributes, **syntactic proficiency, domain-specific linguistic precision,** and **cultural competence,** were added to the inventory for the human evaluators to assess the model's ability to generate language as native Arabic clinicians do. The attributes evaluated in this modified PDQI-9 are detailed in **Table 1.**

| PDQI-9 Attribute | Explanation |
|---|---|
| Accurate | The note does not present incorrect information. |
| Thorough | The information presented is comprehensive and lacks omissions. It contains all information that is relevant to the patient. |
| Useful | The information presented is relevant and provides valuable information for patient management. |
| Organized | The note is formatted in a way that is coherent and easy to comprehend. It helps the reader to understand the patient's story and the management of their clinical case. |
| Comprehensible | The note is straightforward, with no unclear or hard-to-understand sections. |
| Succinct | The note does not contain redundant information and presents relevant information in a concise, direct manner. |
| Synthesized | The note demonstrates the AI's comprehension of the patient's condition and its capability to formulate a care plan. |
| Internally Consistent | The facts presented within the note are consistent with each other and do not contradict the patient's story, each other, or known medical knowledge. |
| Free from Bias | The note is unbiased and includes only information that can be verified by the transcript, without being influenced by the patient's characteristics or the nature of the visit. |
| Free from Hallucinations | The information in the note aligns with the content of the transcript, without any factual inaccuracies or AI-generated hallucinations. |
| **Syntactic Proficiency** | The note correctly uses grammar, relevant vocabulary, and sentence structure to construct accurate sentences as a native speaker would. |
| **Domain-Specific Linguistic Precision** | The note correctly uses medical vocabulary and domain-specific phrasing that best fits the presented clinical scenario. The note utilizes language that native clinicians would use to write their note. |
| **Cultural Competence** | The note's language accurately reflects and respects the cultural, linguistic, and social nuances in Arabic. This includes understanding of culturally specific health beliefs or practices, OR the way that language is utilized surrounding medicine itself. (For example, "palliative care" is not an explicit concept defined in Arabic, or many serious illnesses such as cancers are instead referred to as "the illness.") |

**Table 1.** Items of the modified PDQI-9 for AI-generated summary evaluation.

## Results

| Metric | JAIS | AraSum |
|:---:|:---:|:---:|
| Precision | 0.364 | 0.557 |
| Recall | 0.160 | 0.549 |
| F1 Score | 0.220 | 0.552 |
| ROUGE-1 | 0.000 | 0.452 |
| ROUGE-2 | 0.000 | 0.354 |
| ROUGE-L | 0.000 | 0.452 |
| BLEU | 0.012 | 0.209 |
| BERTScore F1 | 0.724 | 0.807 |

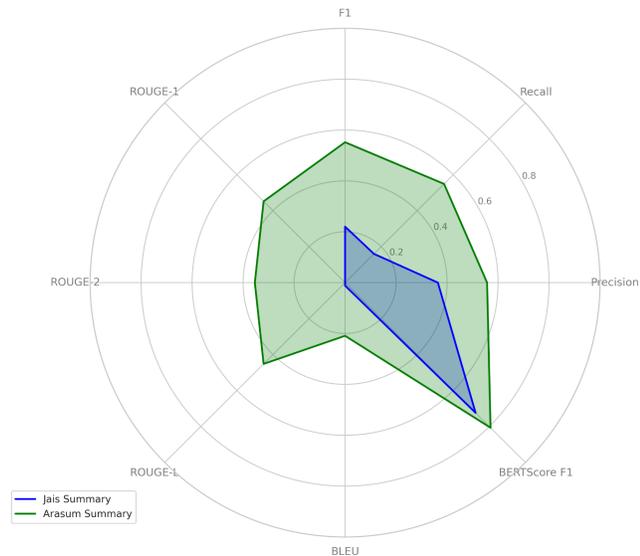

**Figure 1.** AI-centric metrics on summarization performance

**Figure 1** shows the results of the quantitative analysis of AI-centric metrics on summarization performance. Compared to the ground truth, the AI summary by Sporo AraSum far outperforms JAIS in their ability to collect relevant clinical information, synthesize medical knowledge, and turn it into comprehensible summaries. In fact, evaluators also noted from the three clinical vignettes that some AI summaries generated by JAIS were incomplete, indicating that JAIS is unable to handle the volume of medical information necessary for clinical summarization. In contrast, Sporo AraSum's summaries were evaluated both in Arabic and in English translation, and demonstrated far superiority in its ability to create comprehensive and accurate summaries across the board.

**Table 2** gives a representative evaluation of the summaries created by Sporo AraSum vs. JAIS. Both models performed exceptionally well in their ability to synthesize language in the summaries, but the scoring of clinical attributes demonstrate large differences in clinical summarization capabilities. This preliminary finding demonstrates that for domain-specific purposes in clinical workflow, Sporo AraSum is a superior model to be implemented in the clinic than JAIS.

| PDQI-9 Attribute | Summary 1 (AraSum) | Summary 2 (JAIS) |
|---|---|---|
| Accurate | 5 | 5 |
| Thorough | 5 | 3 |
| Useful | 5 | 4 |
| Organized | 5 | 3 |
| Comprehensible | 5 | 4 |
| Succinct | 5 | 5 |
| Synthesized | 5 | 4 |
| Internally Consistent | 5 | 4 |
| Free from Bias | 5 | 5 |
| Free from Hallucinations | 5 | 5 |
| Syntactic Proficiency | 5 | 5 |
| Domain-Specific Linguistic Precision | 5 | 5 |
| Cultural Competence | 5 | 5 |

**Table 2.** Representative evaluation of AI-generated clinical summaries

## Discussion

This case study highlights Sporo AraSum's superior capabilities in Arabic clinical documentation and workflow management compared to JAIS, particularly in the context of medical summarization and decision-making support. By addressing the linguistic, cultural, and technical challenges inherent to the Arabic language, Sporo AraSum demonstrates significant advancements in precision, contextual understanding, and clinical utility, setting a new benchmark for AI models in multilingual healthcare applications.

*Addressing Linguistic Complexity*
Arabic's linguistic intricacies, including its rich morphological structure and diglossia, pose challenges that many existing AI models struggle to overcome. JAIS, despite being a leading Arabic NLP model, demonstrates limitations in handling domain-specific medical terminology and the nuanced syntax required for clinical documentation. Sporo AraSum's ability to accurately synthesize and contextualize medical information reflects its robust architecture and specialized training tailored to Arabic's complexities. The superior F1 scores and evaluator feedback indicate Sporo AraSum's proficiency in capturing subtle linguistic variations and domain-specific vocabulary, ensuring accurate and comprehensive summaries.

*Clinical Precision and Workflow Integration*
Sporo AraSum consistently outperformed JAIS in metrics such as clinical content recall, precision, and organization, as well as qualitative attributes like thoroughness and usefulness. These metrics are vital for ensuring that clinical documentation aids in effective patient care and decision-making. The modified PDQI-9 scores further underscore AraSum's ability to generate concise, accurate, and culturally competent summaries. For instance, its superior performance in domain-specific linguistic precision and cultural competence demonstrates an understanding of not only medical terminology but also the culturally sensitive language necessary in Arabic-speaking regions. This is particularly crucial for fostering trust and understanding in patient-provider interactions.

*Mitigating Risks of AI Hallucination*
One of the critical concerns in AI-generated documentation is the risk of introducing hallucinations or inaccuracies. Both models scored well in this attribute, but Sporo AraSum's consistently high accuracy and internal consistency across all test scenarios highlight its reliability. This suggests that Sporo AraSum's advanced training methodology minimizes the inclusion of irrelevant or erroneous information, a key factor in mitigating risks in clinical decision-making.

*Utility in Multilingual and Cross-Cultural Healthcare*
The findings from this study are not only relevant to Arabic-speaking regions but also have broader implications for the development of multilingual AI models in healthcare. The methodologies behind Sporo AraSum's creation serves as a blueprint for creating language-specific models that address the unique linguistic and cultural challenges of underrepresented languages. Such advancements are critical in bridging communication gaps in diverse healthcare settings, enabling better patient outcomes and reducing disparities in care delivery.

*Limitations and Future Directions*

While Sporo AraSum demonstrates significant improvements over JAIS, this study relies heavily on synthetic data due to the scarcity of real-world clinical datasets in Arabic. While synthetic data has proven effective for evaluation, future research should incorporate real-world clinical conversation datasets to validate these findings further. Additionally, expanding evaluations to include a wider variety of clinical scenarios and integrating user feedback from Arabic-speaking healthcare professionals would provide a more comprehensive assessment of the model's practical utility.

Further exploration is also needed into the integration of Sporo AraSum into clinical workflows, particularly in developing seamless interfaces for real-time documentation and decision support. Finally, expanding the model's capabilities to encompass other dialects and regional variations of Arabic could enhance its applicability across different Arabic-speaking populations.

*Conclusion*

Sporo Health has produced a track history of outperforming foundational models in domain-specific, healthcare-centric tasks through prior case studies.[13],[14] Sporo AraSum emerged as a powerful tool for Arabic clinical documentation, surpassing the current leading model, JAIS, in addressing linguistic and domain-specific challenges. Its ability to generate precise, culturally competent, and clinically useful summaries positions it as a pivotal solution for improving patient care in Arabic-speaking regions. This study underscores the potential of tailored AI models to transform multilingual healthcare by bridging language barriers, enhancing communication, and optimizing clinical workflows.